\documentclass{article}
\usepackage{ijcai26}

\usepackage{times}
\usepackage{soul}
\usepackage{url}
\usepackage[hidelinks]{hyperref}
\usepackage[utf8]{inputenc}
\usepackage[small]{caption}
\usepackage{graphicx}
\usepackage{amsmath}
\usepackage{amsthm}
\usepackage{booktabs}
\usepackage{algorithm}
\usepackage{algorithmic}
\usepackage[switch]{lineno}
\usepackage{verbatim}
\usepackage{colortbl}
\usepackage{mdframed}
\usepackage{adjustbox}
\usepackage{tabularx}
\usepackage{enumitem}
\usepackage{tcolorbox}
\usepackage{natbib}
\usepackage{blindtext}
\usepackage{multirow}
\usepackage[linesnumbered,ruled,vlined,algo2e]{algorithm2e}
\definecolor{shadecolor}{gray}{.9}
\definecolor{Gray}{gray}{0.9}
\tcbuselibrary{skins, breakable}  

\newcommand{\our}[1]{\textsc{Plan-MCTS}}

\title{\our{}: Plan Exploration for Action Exploitation in Web Navigation}

\author{
Weiming Zhang$^1$\thanks{Work done during an internship at OPPO.}
\and
Jihong Wang$^2$
\and
Jiamu Zhou$^2$
\and
Qingyao Li$^1$
\and
Xinbei Ma$^1$
\and
\\
Congmin Zheng$^1$
\and
Xingyu Lou$^2$\thanks{Corresponding author.}
\and
Weiwen Liu$^1$\footnotemark[2]
\and
Zhuosheng Zhang$^1$
\and
\\
Zhaoxiang Wang$^2$
\and
Jun Wang$^2$
\and
Yong Yu$^1$
\and
Weinan Zhang$^1$\footnotemark[2]
\\
\affiliations
$^1$Shanghai Jiao Tong University\\
$^2$OPPO Research Institute\\
\emails
\{WeimingZhang\_2020, wwliu, wnzhang\}@sjtu.edu.cn,
louxingyu@oppo.com,
}

\begin{document}

\maketitle

\begin{abstract}
Large Language Models (LLMs) have empowered autonomous agents to handle complex web navigation tasks. While recent studies integrate tree search to enhance long-horizon reasoning, applying these algorithms in web navigation faces two critical challenges: sparse valid paths that lead to inefficient exploration, and a noisy context that dilutes accurate state perception. To address this, we introduce \our{}, a framework that reformulates web navigation by shifting exploration to a semantic Plan Space. By decoupling strategic planning from execution grounding, it transforms sparse action space into a Dense Plan Tree for efficient exploration, and distills noisy contexts into an Abstracted Semantic History for precise state awareness. To ensure efficiency and robustness, \our{} incorporates a Dual-Gating Reward to strictly validate both physical executability and strategic alignment and Structural Refinement for on-policy repair of failed subplans. Extensive experiments on WebArena demonstrate that \our{} achieves state-of-the-art performance, surpassing current approaches with higher task effectiveness and search efficiency.

\end{abstract}

\section{Introduction}
Powered by recent breakthroughs in Large Language Models (LLMs), autonomous agents have shown promising intelligence in web navigation, significantly alleviating human labor across tasks such as travel planning, online shopping, and research~\citep {wang2024survey,guo2024large,xi2025rise}. As the primary interface for digital interaction, the web features an environment with massive interactive components. Navigating such a rich and complex space requires robust perception and reasoning capabilities, making web navigation a critical testbed and research focus for agent intelligence.

Inspired by the recent success of scaling test-time compute for complex reasoning~\citep{jaech2024openai,qin2024o1,zhang2025survey}, researchers have actively integrated search algorithms into autonomous agent systems. This paradigm empowers agents to perform iterative trial-and-error reasoning, exploring diverse solution paths to tackle complex tasks. For instance, Tree of Thoughts~\citep{yao2023tree} and Search Agents~\citep{koh2024tree} employ Best-First Search to augment decision-making, while \citet{yu2024exact} and \citet{zhang2025webpilot} further utilize Monte Carlo Tree Search (MCTS) to balance exploration and exploitation~\citep{zhou2023language}. Allowed to explore diverse possibilities, agents can recover from dead ends and identify the potentially optimal path to the goal.

\begin{figure}[t]
    \centering
    \includegraphics[width=1.0\linewidth]{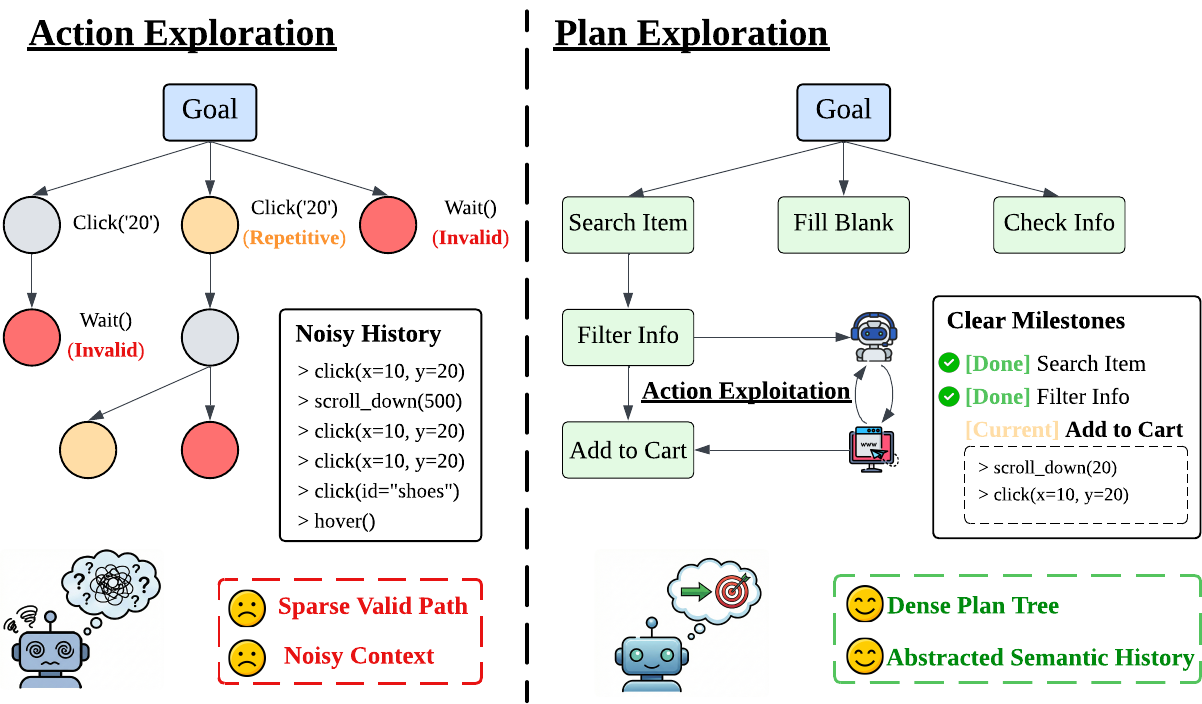}
    \caption{Comparison between Action Space Search and Plan Space Search in web navigation.} 
    \label{fig: motivation} \vspace{-10pt}
\end{figure}

Despite this progress, applying search algorithms in web navigation faces two significant challenges, as depicted in Figure~\ref{fig: motivation}: (1) \textbf{Sparse Valid Path.} Web navigation typically presents a massive action space due to the abundance of interactive elements and distractors. However, only a small subset of these actions leads to meaningful progress. Consequently, naive exploration in this raw space yields a large number of erroneous trajectories, leading the agent to waste its limited budget on fruitless exploration. We define this structural limitation as sparse valid paths, where valid solutions are embedded sparsely within a vast combinatorial space of dead ends. (2) \textbf{Noisy Context.} Atomic execution inherently accumulates a long historical context, characterized by verbose execution traces such as precise coordinates and scroll events. Although capturing the precise mechanics of interaction, they suffer from low information density. This unstructured noise dilutes the agent's perception of task progress, burying critical milestones under excessive low-level details. Consequently, the agent struggles to maintain precise state awareness and accurately ground subsequent commands.

To address these limitations, we propose shifting the search granularity in web navigation to \textbf{Plan Space}. In this space, the agent reasons over high-level natural language intents rather than getting lost in raw atomic actions. Realizing this paradigm, we introduce \our{}, a framework that reformulates web navigation by decoupling strategic planning from execution grounding. As illustrated in Figure~\ref{fig: motivation}, this paradigm resolves the aforementioned challenges from two aspects: (1) \textbf{Dense Plan Tree.} We construct a shallow search tree where each edge represents a high-utility subplan, abstracting sparse action chains into a structured hierarchy of meaningful intents for efficient exploration. (2) \textbf{Abstracted Semantic History.} We maintain a context of verified milestones, distilling the verbose stream of atomic traces into a coherent narrative of task progress for maintaining precise state awareness.

To make \our{} efficient and effective in practice, we incorporate two complementary mechanisms into the MCTS process. First, we employ \textbf{Dual-Gating Evaluation}, which strictly validates both the physical completion of the subplan and its strategic contribution to the global goal. Second, we introduce \textbf{Structural Refinement}, an on-policy repair mechanism that dynamically adjusts the content of a subplan when execution falters, rather than simply discarding the path. Together, these mechanisms ensure that the search is driven by verified, high-utility progress signals while empowering the agent to recover from local execution errors, enabling efficient convergence within a limited budget.

In summary, our contributions are listed as follows:
\begin{itemize}
    \item We propose \our{}, a novel framework that explores high-level plans and executes low-level actions. To the best of our knowledge, we are the first to conduct tree search within the high-level plan space in autonomous web navigation.
    \item We introduce Structural Refinement and Dual-Gating Reward mechanisms that make our model more stable and efficient. They synergize to regulate the search dynamics, balancing exploration costs against decision quality to discover optimal execution trajectories.
    \item We demonstrate that \our{} achieves state-of-the-art performance on the WebArena benchmark. Empirical results confirm its superiority over strong baselines, achieving significantly higher task effectiveness and search efficiency.
\end{itemize}

\section{Related Works}
\subsection{Autonomous Web Agents}
Web navigation tasks require agents to execute natural language instructions across dynamic web interfaces to achieve multi-step objectives~\citep{zhou2023webarena,ning2025survey,song2025colorbench,li2025coloragent}. Recent LLM-based approaches perceive these environments via HTML or screenshots, typically adopting a linear execution paradigm to optimize specific capabilities~\citep{zheng2024gpt,he2024webvoyager}. For example, ReAct~\citep{yao2022react} and SteP~\citep{sodhi2023step} enhance reasoning through "think-before-act" cycles and task decomposition, while Agent Workflow Memory~\citep{wang2024agent} and AgentOccam~\citep{yang2024agentoccam} improve information processing via experience retrieval and context pruning, respectively. However, due to their strict linearity, these methods remain vulnerable to error accumulation and repetitive loops, lacking the mechanism to backtrack or recover from local optima. As LLM reasoning capabilities scale with increased inference budgets~\citep{balachandran2025inference}, recent research has integrated search algorithms to mitigate the error propagation inherent in linear methods. Inspired by Tree of Thoughts~\citep{yao2023tree} and Tree Search for LM Agents~\citep{koh2024tree}, works such as WebPilot~\citep{zhang2025webpilot} and R-MCTS~\citep{yu2024exact} utilize Monte Carlo Tree Search (MCTS)~\citep{zhou2023language} to balance exploration and exploitation. Branch-and-Browse~\citep{he2025branch} and WebOperator~\citep{dihan2025weboperator} further enhance search robustness and efficiency in dynamic environments. However, these methods predominantly operate on the atomic action space, generating noisy, low-level execution traces that lack semantic abstraction and obscure the perception of task progress in long-horizon scenarios.

\subsection{Search-Based Decision Making for LLMs}
Recent research leverages increased inference budgets to enhance problem-solving via tree search algorithms. Tree of Thoughts~\citep{yao2023tree} generalizes chain-of-thought reasoning into tree structures, while LATS~\citep{zhou2023language} integrates Monte Carlo Tree Search (MCTS) for self-reflective trajectory optimization. This paradigm has proven effective across diverse NLP tasks, including Question Answering~\citep{schimanski2024towards}, Mathematical Reasoning~\citep{zhang2024accessing}, and Code Generation ~\citep{feng2023alphazero}, where search effectively navigates complex solution spaces. Building on this, LLM-based agents have incorporated MCTS to bolster decision-making capabilities~\citep{swiechowski2023monte}. However, searching directly over vast atomic action spaces often encounters severe sparsity, leading to incoherent state transitions and inefficient exploration. Inspired by PlanSearch~\citep{wang2024planning} and RethinkMCTS~\citep{li2025rethinkmcts}, we recognize that LLMs exhibit superior reasoning in the natural language space compared to concrete downstream action spaces. Searching within this semantic space preserves logical coherence and significantly enhances efficiency. This insight motivates us to construct the first web agent framework that performs adaptive search directly over natural language subplans.

\begin{figure*}[t]
    \centering
    \includegraphics[width=1.0\textwidth]{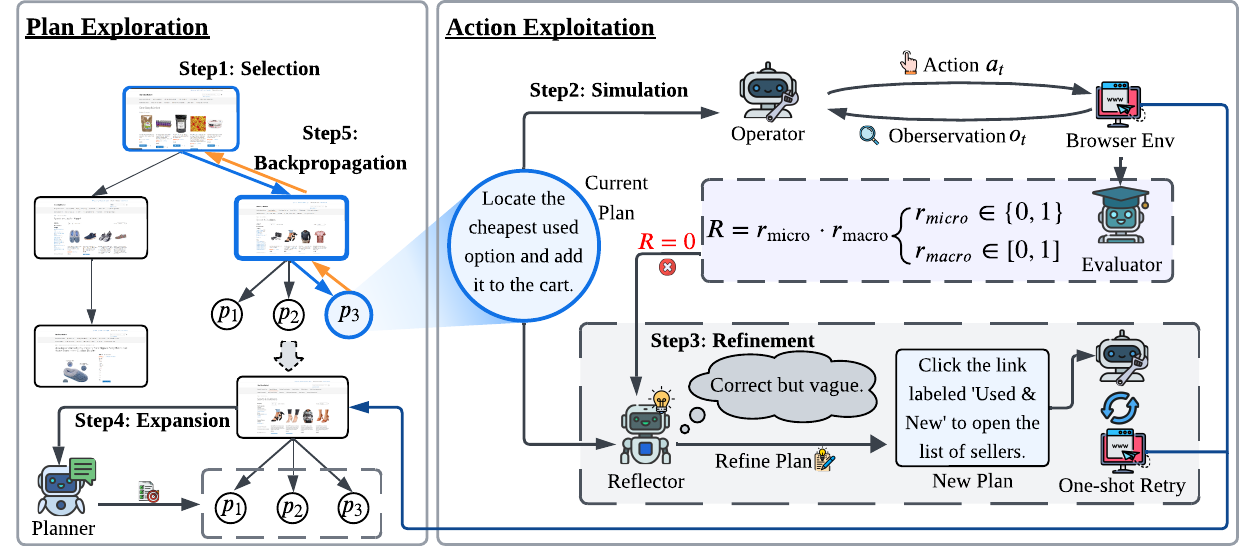}
    \caption{The overall framework of \our{}. We employ MCTS to explore high-level plans for strategic reasoning, while utilizing an Operator to exploit these plans through atomic action grounding.}
    \label{fig: framework}\vspace{-10pt}
\end{figure*}

\section{Preliminaries}
\subsection{Problem Formulation}
We formulate the autonomous web navigation task as a Partially Observable Markov Decision Process (POMDP), defined by the tuple $\mathcal{M} = (\mathcal{S}, \mathcal{A}, \mathcal{O}, \mathcal{T}, \mathcal{R}, \mathcal{G})$. The interaction proceeds in discrete time steps $t=0, 1, \dots, T$. 

Specifically, the process is initiated by a natural language instruction $g \in \mathcal{G}$ that specifies the user's intent. At each step, the environment resides in a latent state $s_t \in \mathcal{S}$ (e.g., the underlying Document Object Model (DOM), a hierarchical tree-like representation of the rendered webpage), from which the agent perceives a multi-modal observation $o_t \in \mathcal{O}$. To navigate this environment, the agent samples an atomic action $a_t \in \mathcal{A}$ from a policy $\pi(a_t \mid o_t, H_t, g)$ conditioned on the interaction history $H_t = (o_0, a_0, \dots, a_{t-1}, o_t)$. Each action triggers a state transition $s_{t+1} \sim \mathcal{T}(\cdot \mid s_t, a_t)$, continuing until a termination condition is met, whereupon a sparse binary reward $r \in \{0, 1\}$ is assigned to indicate task success or failure. The agent's objective is to determine an optimal trajectory that maximizes the expected cumulative reward within the step constraints.

\subsection{Monte-Carlo Tree Search}
Monte-Carlo Tree Search (MCTS) is a best-first search algorithm that iteratively builds a decision tree to estimate the optimal policy. The process consists of four phases: \textit{Selection}, \textit{Expansion}, \textit{Simulation}, and \textit{Backpropagation}. During the \textit{Selection} phase, the algorithm traverses from the root to a leaf node by iteratively choosing the child that maximizes the Upper Confidence Bound for Trees (UCT) criterion:
\begin{equation}
    a^* = \operatorname*{arg\,max}_{a \in \mathcal{A}(s)} \left( Q(s, a) + c \cdot \sqrt{\frac{\ln N(s)}{N(s, a)}} \right)
\end{equation}
where $Q(s, a)$ denotes the estimated value (expected reward) of taking action $a$ in state $s$, while $N(s)$ and $N(s, a)$ represent the visit counts of the parent node and the specific action edge, respectively. The exploration constant $c$ modulates the trade-off between exploiting high-value paths and exploring less-visited alternatives.

In contrast to standard MCTS, which operates directly on the atomic action space $\mathcal{A}$, we propose lifting the search process to a high-level Semantic Plan Space. We formally define this methodology in the following section.

\section{Methodology}
\subsection{Framework Overview}
\label{sec: overview}
In this section, we introduce \textbf{\our{}}, a framework that reformulates web navigation as a search problem over a semantic plan space.

The complex nature of web environments makes the random Simulation of standard MCTS impractical. To address this, we fundamentally redefine the simulation phase to suit the web context. Rather than employing stochastic rollouts to estimate future rewards, our Simulation focuses on execution and evaluation: the selected subplan is physically grounded by the Operator to drive state transitions, enabling a rigorous assessment of actual execution quality.

To minimize the high latency costs associated with such grounded executions, we organize the overall process into five phases: (1) \textit{Selection} of a promising subplan based on priors; (2) \textit{Simulation}, where the subplan is executed via atomic actions in the environment and subsequently evaluated; (3) \textit{Refinement}, an step to repair failed trajectories dynamically; (4) \textit{Expansion}, where the planner generates the next set of candidate subplans based on the newly verified state; and (5) \textit{Backpropagation}, which updates the search tree statistics with the computed rewards.

Three key features characterize our design:
\begin{itemize} \item \textbf{Plan Exploration for Action Exploitation:} We decouple high-level intent generation from low-level action grounding. The search algorithm explores the semantic Plan Space to determine optimal intents, while a local operator efficiently exploits these intents by grounding them into atomic actions.

\item \textbf{Dual Evaluation:} To ensure reliable value estimation, we employ a Dual-Gating mechanism that combines binary \textit{micro-verification} (subplan completion) with scalar \textit{macro-assessment} (global progress).

\item \textbf{Structural Refinement:} To ensure robustness, we introduce a dynamic refinement mechanism. When execution fails, the system diagnoses the error and modifies the subplan to recover from local execution noise.
\end{itemize}

\subsection{Selection}
In plan space, we construct a search tree where each node corresponds to a verified deterministic state $s$. The root represents the initial state $s_0$. Crucially, unlike standard formulations where edges imply atomic actions, each directed edge stemming from state $s$ corresponds to a subplan $p \in \mathcal{P}$ (e.g., ``Filter results by price''). For every edge $(s, p)$, we maintain two critical statistics: the \textit{Q-Value} $Q(s, p)$, representing the estimated quality of the subplan, and the \textit{Visit Count} $N(s, p)$, tracking how often this intent has been selected.

The Selection phase navigates this structure from the root $s_0$ to identify a promising leaf for subsequent expansion. At each depth, the algorithm resolves the exploration-exploitation dilemma by selecting the optimal subplan $p^*$ that maximizes the UCT criterion:

\begin{equation} p^* = \operatorname*{arg,max}_{p \in \mathcal{P}(s)} \left( Q(s, p) + c \cdot \sqrt{\frac{\ln N(s)}{N(s, p)}} \right) \end{equation}
where $\mathcal{P}(s)$ denotes the set of candidate subplans available at state $s$, and $N(s)$ represents the total visit count of the current state. Crucially, unlike standard approaches that strictly select nodes, our selection process targets edge $p^*$ and identifies the specific high-level intent that requires physical execution in the subsequent Simulation phase.

\subsection{Simulation}
\label{sec:simulation}

As discussed in Sec.~\ref{sec: overview}, to accommodate the irreversible nature of web environments, we replace stochastic rollouts with an execution-and-evaluation process. This phase consists of two sequential stages: \textit{Action Grounding} and \textit{Dual-Gating Evaluation}.

\paragraph{Action Grounding}
Upon selecting the high-level intent $p^*$, the Operator(parameterized as $\pi_{\text{op}}$) executes the selected subplan $p$. The Operator focuses strictly on local execution details. It maintains a hierarchical context comprising the global high-level plan history $H_{\mathcal{P}}$ and the \textit{local} atomic action history $h_{\text{loc}}$ specific to the current subplan. This design prevents the agent's context window from being saturated by the verbose action traces of previous subplans.

At each time step $t$, the Operator generates an atomic action $a_t$ based on the current observation $o_t$ and the hierarchical context:
\begin{equation}
    a_t \sim \pi_{\text{op}}(a_t \mid o_t, H_{\mathcal{P}}, h_{\text{loc}}, p)
\end{equation}

The environment then transitions to the next state $s_{t+1}$. This iterative grounding continues until the subplan is fulfilled or a termination condition is met, producing a concrete trajectory segment for subsequent evaluation.

\paragraph{Dual-Gating Evaluation}
Upon reaching the new state $s'$ following the execution of subplan $p$, we employ a Dual-Gating Evaluation mechanism to assess the trajectory quality. This mechanism comprises two hierarchical components:
\begin{itemize}
    \item \textbf{Micro-Score ($r_{\text{micro}} \in \{0, 1\}$):} A binary metric strictly verifying if the executed actions successfully fulfilled the subplan's intent. Any execution failure or semantic deviation results in $r_{\text{micro}}=0$.
    \item \textbf{Macro-Score ($r_{\text{macro}} \in [0, 1]$):} A scalar metric estimating global progress towards the instruction $g$. To mitigate assessment variance, we average scores from $N$ independent evaluations: $r_{\text{macro}} = \frac{1}{N} \sum_{i=1}^{N} \text{Score}_i(s', g)$.
\end{itemize}
The final reward is computed as the product $R(s, p) = r_{\text{micro}} \cdot r_{\text{macro}}$. This gating logic ensures that the agent receives high rewards only when it executes feasible plans that are strategically aligned with the ultimate goal.

\subsection{Refinement}
\label{sec:refinement}
This phase serves as an on-policy repair mechanism. If the Action Grounding phase fails to verify the subplan (i.e., $r_{\text{micro}}=0$), the Reflector Agent (parameterized as $\pi_{\text{ref}}$) is triggered to perform \textbf{Structural Refinement}.

Rather than immediately treating the execution failure as a terminal signal, the Reflector analyzes the failure context to diagnose the root cause. It then generates a revised subplan $p'$ conditioned on the current observation $o_t$ and the high-level plan history $H_{\mathcal{P}}$:

\begin{equation}p' \sim \pi_{\text{ref}}(p' \mid o_t, H_{\mathcal{P}}, p)\end{equation}

This rectified instruction $p'$ takes the form of either a finer-grained decomposition or a corrected directive, specifically designed to overcome the current execution failure. By dynamically adjusting the plan, the Reflector ensures that the search is robust to local execution noise and prevents valid plans from being rejected due to temporary execution issues.

\begin{table*}[]
\tiny
\resizebox{\textwidth}{!}{%
\begin{tabular}{llllllll}

\toprule
\textbf{Method}          & \textbf{Model}      & \textbf{Shopping} & \textbf{CMS} & \textbf{Gitlab} & \textbf{Map}   & \textbf{Reddit} & \textbf{Average} \\ \midrule
Webarena~\citep{zhou2023webarena}          & GPT-4-Turbo & 16.6     & 15.9           & 10.0   & 22.9  & 21.7   & 16.5   \\
BrowserGym~\citep{chezelles2024browsergym}       & GPT-4o      & 26.6     & 28.0             & 21.4   & 18.4  & 22.8   & 24.3   \\ \midrule
AutoEval~\citep{pan2024autonomous}        & GPT-4       & 39.6     & 20.9           & 25.0   & 27.5  & 20.8   & 26.9   \\
SteP~\citep{sodhi2023step}              & GPT-4-Turbo & 33.2     & 32.4           & 26.7   & 35.8  & 52.8   & 33.3   \\
AWM~\citep{wang2024agent}               & GPT-4-Turbo & 32.1     & 29.1           & 35.0   & 42.2  & 54.7   & 35.5   \\
CER~\citep{liu2025contextual}               & GPT-4o      & 32.8     & 41.2           & 37.2   & 30.4  & 41.2   & 36.7   \\
AgentOccam~\citep{yang2024agentoccam}        & GPT-4-Turbo & 43.3     & 46.2           & 38.9   & 52.3  & 67.0   & 45.7   \\
Search Agent~\citep{koh2024tree}      & GPT-4o      & 27.8     & 16.5           & 13.9   & 26.6  & 11.3   & 19.2   \\
WebPilot~\citep{zhang2025webpilot}          & GPT-4o      & 36.9     & 24.7           & 39.4   & 33.9  & 64.1   & 37.2   \\
Branch-and-Browse~\citep{he2025branch} & GPT-4o      & 34.6     & 26.4           & 36.7   & 46.8  & 50.9   & 35.8   \\ \rowcolor{Gray}
\our{}         & GPT-4o      & 43.6     & 40.5          & 33.7  & 24.6 & 54.1     & 39.2   \\ \bottomrule
\end{tabular}%
}
\caption{Performance of \our{} and other baselines on WebArena. We report the Success Rate (\%) across five distinct domains.}
\label{tab:main_results} \vspace{-10pt}
\end{table*}

\subsection{Expansion}
\label{sec:expansion}
Once a valid state transition to $s'$ is confirmed via either the Simulation phase or a successful Refinement, the algorithm proceeds to expand the search frontier. This step is managed by the Planner(parameterized as $\pi_{\text{plan}}$).

Given the new verified state $s'$ and the cumulative global plan history $H_{\mathcal{P}}$, the Planner generates a set of $k$ candidate subplans for the subsequent step:

\begin{equation}{p_1, p_2, \dots, p_k} \sim \pi_{\text{plan}}(\cdot \mid s', H_{\mathcal{P}})\end{equation}
These candidates are instantiated as new child edges stemming from node $s'$ in the search tree. Crucially, $H_{\mathcal{P}}$ records the sequence of previously executed subplans rather than verbose atomic actions, thereby filtering out low-level execution noise.

\subsection{Backpropagation}
In the final phase of the iteration, the reward signal $R$ derived from the Simulation (or Refinement) phase is propagated backward to update the search tree statistics. Starting from the current leaf node and traversing up to the root $s_0$, we update the statistics for every subplan edge $(s, p)$ along the selected path:
\begin{align}
    N(s, p) &\leftarrow N(s, p) + 1 \\
    Q(s, p) &\leftarrow Q(s, p) + \frac{R - Q(s, p)}{N(s, p)}
\end{align}
Here, $N(s, p)$ is incremented to record the visit, and $Q(s, p)$ is updated incrementally to reflect the new mean value. This mechanism ensures that the estimated utility of each subplan progressively converges to its true expected value, thereby guiding future Selection phases toward more robust and high-value strategies.

\section{Experiments}
In this section, we structure our experiments to answer the following research questions:

\noindent \textbf{RQ1 (Effectiveness)}: How does \our{} perform against the baselines, especially those using atomic action search?

\noindent \textbf{RQ2 (Efficiency)}: Does searching in the plan space yield superior efficiency compared to the action space?

\noindent \textbf{RQ3 (Scalability)}: Does the \our{} exhibit favorable scaling laws with increased inference compute budgets?

\noindent \textbf{RQ4 (Context)}: How does decoupling global plan history from local action traces mitigate context pollution?

\noindent \textbf{RQ5 (Ablation)}: What are the individual contributions of the Dual-Gating Reward and Structural Refinement mechanisms?


\subsection{Experimental Settings}
\paragraph{Benchmarks}
We conduct our evaluation on \textbf{WebArena}~\citep{zhou2023webarena}, a highly realistic and reproducible web environment designed to assess autonomous multimodal agents on complex, long-horizon tasks. 
The benchmark comprises 812 distinct tasks instantiated from 241 task templates, spanning five diverse domains: \textit{Shopping, Shopping Admin, Reddit, Map,} and \textit{Gitlab}. 
This extensive coverage and environmental fidelity pose significant challenges to the agent's capabilities in sequential decision-making, long-term planning, and robustness to noise. 
For evaluation, WebArena employs a rigorous execution-based mechanism to determine task correctness upon completion. Accordingly, we adopt the \textbf{Success Rate (SR)} as our primary evaluation metric.

\paragraph{Baselines}
We compare \our{} against a comprehensive set of baselines, categorized into two distinct paradigms:
(1) \textbf{Sequential Agents}: This category includes vanilla agents operating within a single reasoning-acting loop without backtracking, such as the standard ReAct~\citep{yao2022react} implementation and BrowserGym~\citep{chezelles2024browsergym}.
(2) \textbf{Policy-Based Strategies}: This category encompasses advanced frameworks ranging from linear enhancements (AutoEval~\citep{pan2024autonomous}, SteP~\citep{sodhi2023step}, AWM~\citep{wang2024agent}, and AgentOccam~\citep{yang2024agentoccam}) to search-based methods. For the latter, we compare against WebPilot~\citep{zhang2025webpilot}, Branch-and-Browse~\citep{he2025branch}. Detailed configurations for all baselines are provided in the Appendix.

\paragraph{Implementation Details}
We build our framework upon \textbf{BrowserGym}~\citep{chezelles2024browsergym}, leveraging Playwright for programmatic web interaction. 
Our agents utilize multimodal observations, integrating Set-of-Marks (SoM)~\citep{yang2023set} and Accessibility Trees (AxTree)~\citep{zhou2023webarena}, with refined action spaces adapted from AgentOccam~\citep{yang2024agentoccam} to minimize execution errors.

Crucially, to ensure fairness, all self-implemented baselines (i.e., Action Search and Action MCTS) operate within this \textit{identical} environment, sharing the same input prompting and optimization protocols.
We configure the search parameters with a budget of $N=10$, maximum depth $d=5$, and width $k=3$. 
More details are provided in the Appendix.

\begin{table*}[t]
\centering
\tiny
\resizebox{0.9\textwidth}{!}{%
\begin{tabular}{llllllll}
\toprule
\textbf{Model}                & \textbf{Method} & \textbf{Shopping} & \textbf{CMS} & \textbf{Gitlab} & \textbf{Map} & \textbf{Reddit} & \textbf{Average} \\ \hline
\multirow{4}{*}{Qwen3-VL-32B} & Action Search   & 33.2              & 21.1         & 35.1            & 21.1         & 33.6            & 29.3            \\
                            & Action MCTS   & 37.2 & 26.4 & 33.6 & 20.9 & 34.4 & 29.3 \\
                            & Plan Search   & 36.7 & 31.2 & 36.9 & 24.8 & 43.9 & 34.8 \\ \rowcolor{Gray}
                            & \our{}     & 42.4 & 37.6 & 38.5 & 23.8 & 50.7 & 38.8 \\ \midrule
\multirow{4}{*}{GPT-5-mini} & Action Search & 46.4 & 54.0  &  43.8 & 28.0 & 51.3 & 45.5  \\
                            & Action MCTS   & 46.5 & 53.7 & 44.5  &  23.5 & 57.1 & 45.8 \\
                            & Plan Search   & 49.1 & 55.4 & 49.4 & 29.1 & 65.1 & 50.0 \\ \rowcolor{Gray}
                            & \our{}     & 57.8 & 62.7 & 56.3 & 26.1 & 68.1 & 55.3 \\ \bottomrule
\end{tabular}%
}
\caption{Performance comparison between Action-Space and Plan-Space methods on  WebArena across different backbones. All methods are implemented within an identical environment to ensure a fair comparison.}
\label{tab:plan} 
\end{table*}

\subsection{Overall Performance (RQ1)}
We conduct comprehensive evaluations on the full WebArena benchmark. Crucially, to ensure a strictly fair comparison, we implemented both action-level and plan-level variants of Beam Search and MCTS within an identical environment. Our findings can be summarized into two key observations:

(1) \our{} achieves the best overall performance on the WebArena benchmark as Table~\ref{tab:main_results} shows. Notably, under the identical GPT-4o setting, \our{} establishes a lead over prior approaches. This empirical evidence confirms its superior capability in handling complex, long-horizon tasks compared to previous methods.

(2) Methods operating in the Plan Space consistently outperform Action Space variants on both Qwen3-VL-32B and GPT-5-mini, with \our{} achieving the highest success rates. This gap indicates that searching over plans effectively bypasses atomic noise and sparsity, yielding more effective trajectories than the raw action space. Furthermore, the superiority of \our{} over standard Plan Search highlights the critical benefit of the UCT algorithm in balancing exploration and exploitation.

\begin{figure}[t]
    \centering
    \includegraphics[width=1.0\columnwidth]{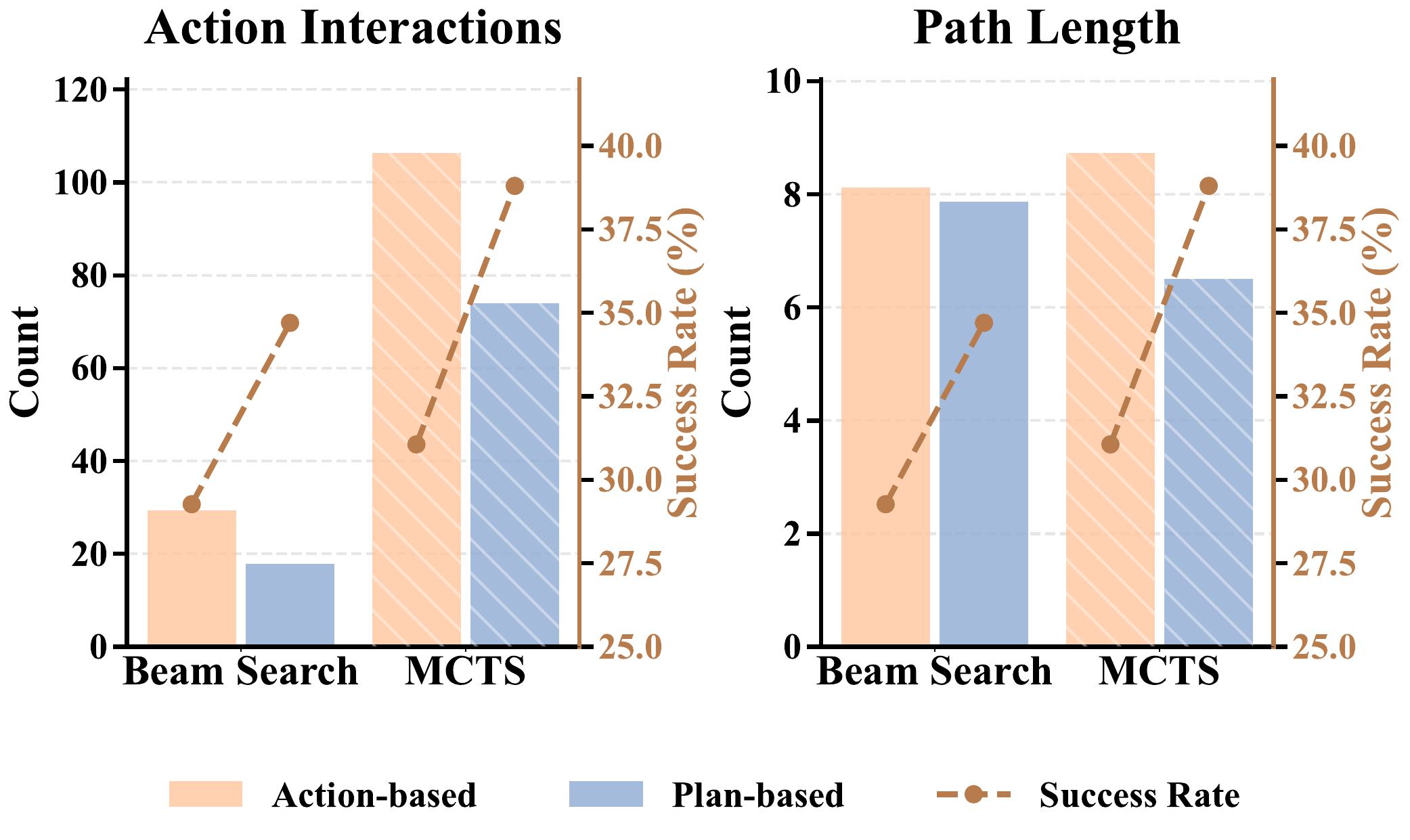}
    \caption{Efficiency comparison between Action-Space and Plan-Space search.} 
    \label{fig: efficiency} \vspace{-5pt}
\end{figure}

\begin{figure}[t]
    \centering
    \includegraphics[width=0.85\linewidth]{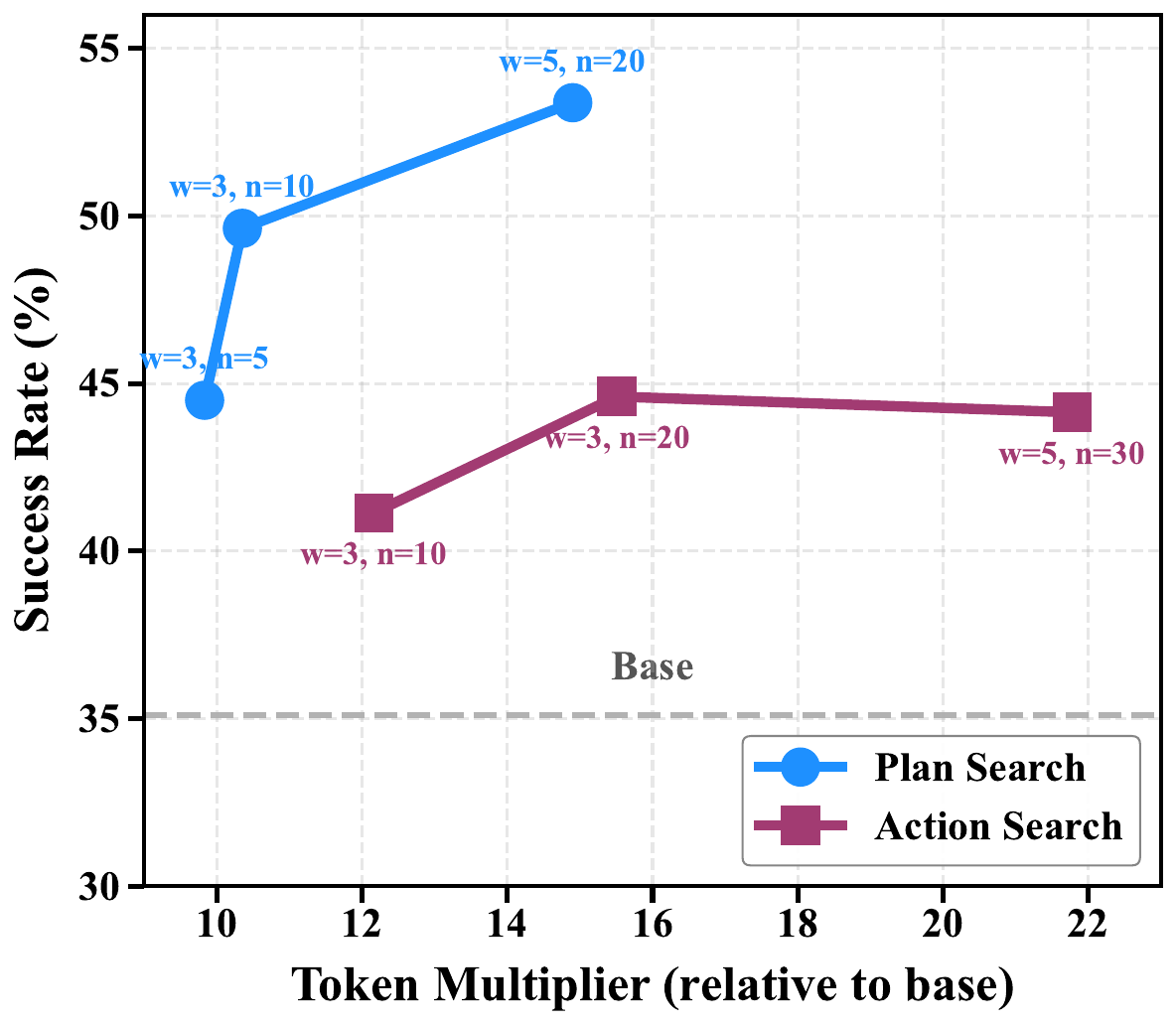}
    \caption{Performance comparison between \our{} and action-level search  when we scale the search budget.} 
    \label{fig: scaling} \vspace{-5pt}
\end{figure}

\subsection{Efficiency Analysis (RQ2)}
We systematically investigate the efficiency and cost advantages of searching within the Plan Space compared to the Action Space. To quantify these benefits, we analyze the search process from two metrics: \textit{Action Interactions} and \textit{Path Length}, while monitoring the Success Rate (SR) to assess the cost-performance trade-off, as illustrated in Figure~\ref{fig: efficiency}.

\paragraph{Plan-Space search demonstrates superior resource efficiency compared to Action-Space search.} Compared to their Action-Space counterparts, Plan Search and \our{} reduce action interactions by 40\% and 28\%, respectively. This substantial reduction validates that the semantic plan space effectively filters out the sparse and noisy branches inherent to atomic interactions, enabling the agent to concentrate the search budget on high-utility strategic moves rather than exhaustive low-level exploration.

\paragraph{Plan-Space search yields structurally more concise and optimal trajectories.} Plan-Space variants consistently achieve shorter path lengths across both search algorithms. We observe a reduction in average path step length compared to Action-Space baselines, indicating that high-level planning effectively prevents the agent from degenerating into aimless exploration or redundant loops.

\subsection{Scaling Analysis (RQ3)}
To investigate the impact of inference compute on performance, we conduct a scaling study by varying the search budget parameters (i.e., width $W$ and limit $N$) on two website domains (\textit{Shopping and \textit{Reddit}}). Crucially, to ensure a fair comparison across disparate search spaces, we align the evaluation metric by total token consumption rather than tree size, mapping the computational cost directly to the Success Rate in Figure~\ref{fig: scaling}. 

\our{} demonstrates superior scalability, converting increased compute into stable performance gains. As illustrated in Figure~\ref{fig: scaling}, we observe a consistent improvement in Success Rate as the search budget and token consumption increase. In contrast, Action Search incurs significantly higher token overhead yet yields negligible growth. This stems from the underlying search structure: the Dense Plan Tree ensures that additional compute is invested in exploring meaningful semantic intents, whereas the inherent sparsity of valid paths in the Action Space renders simple budget scaling ineffective.

\subsection{Impact of Context Decoupling (RQ4)}
\begin{figure}[t]
    \centering
    \includegraphics[width=0.9\linewidth]{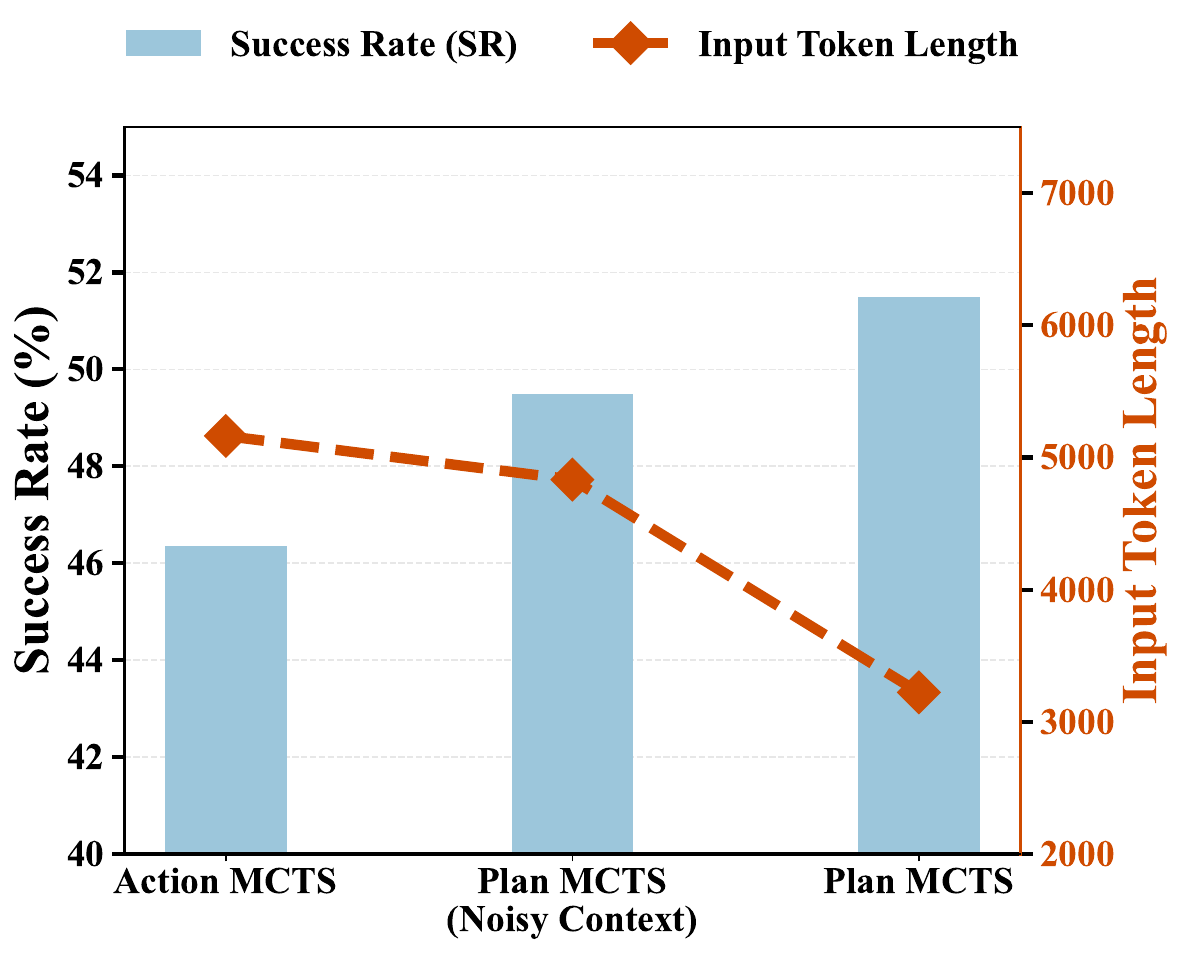}
    \caption{Performance comparison in terms of different context components.} 
    \label{fig: context} \vspace{-5pt}
\end{figure}

We investigate the efficacy of \our{}'s decoupling strategy, which substitutes the noisy atomic context with a clear, plan-level history. We compare this design against two baselines processing the full, undecoupled interaction history: (1) the standard Action-MCTS, and (2) a variant of \our{} operating with raw noisy context. The results in Figure~\ref{fig: context} demonstrate two key benefits:

\begin{itemize}[leftmargin=10pt]
\item \textbf{Superior Efficacy with Reduced Overhead.} Decoupling the context reduces input tokens while simultaneously boosting the Success Rate. Notably, while \our{} with raw history already outperforms Action Search, the decoupled plan-level context further improves performance by effectively filtering out execution noise.
\item \textbf{Enhanced Execution Robustness.} We further evaluate execution stability via the \textit{Subplan Completion Rate}. Empirical results indicate that the noisy context degrades this metric by approximately 10\%, confirming that verbose atomic traces distract the Operator. In contrast, our decoupled context ensures more precise grounding, thereby significantly enhancing execution robustness.
\end{itemize}

\subsection{Ablation Study (RQ5)}
We investigate the individual contributions of the \textbf{Dual-Gating Reward} and \textbf{Structural Refinement} mechanisms in ensuring efficient and stable exploration. We utilize \textit{Budget Utilization} and \textit{Step Length} to quantify their impact on exploration sufficiency and trajectory optimality.
\subsubsection{Reward Design}
We perform an ablation study comparing the Dual-Gating mechanism against two decoupled variants: \textbf{Micro-only} and \textbf{Macro-only} (Table~\ref{tab:ablation}). We draw the following observations:

\begin{itemize}[leftmargin=10pt]
\item \textbf{Micro-only:} Results in low budget utilization but excessive step lengths. The agent gets trapped in local "valid loops," prioritizing executability over global progress.
\item \textbf{Macro-only:} Suffers from inflated budget utilization due to hallucinated success. It wastefully expands invalid branches based solely on intent, neglecting actual plan completion.
\item \textbf{Dual-Gating:} Balances budget utilization while preserving a high Success Rate. By employing executability as a hard gate, it effectively filters execution noise without losing alignment with the global direction.
\end{itemize}

\subsubsection{Refinement Strategy}
To validate the benefits of structural refinement, we compare our \textbf{Refinement} mechanism against a standard \textbf{Reflection} baseline that leverages error experience solely to generate textual feedback without altering the original tree structure.

The results in Table~\ref{tab:ablation} demonstrate that our Structural Refinement significantly outperforms Reflection, achieving a higher Success Rate and more optimal trajectories with reduced search budget. This efficiency stems from the ability to directly modify incorrect subplans, whereas Reflection continuously accumulates error history without rectifying the root cause, leading to inefficient exploration.

\begin{table}[t]
\centering
\scriptsize
\resizebox{\columnwidth}{!}{%

\newcommand{\drop}[1]{\rlap{$_{\textnormal{\tiny (-#1\%)}}$}} 

\begin{tabular}{lccc}
\toprule
\textbf{Method} & \textbf{SR} & \textbf{Budget Util} & \textbf{Step Len} \\ 
\midrule
\textbf{\our{}} & \textbf{51.5} & \textbf{28.3} & \textbf{7.8} \\ 
\midrule
\multicolumn{4}{l}{\textit{Ablation on Reward Design}} \\ 
Micro-only reward & 47.6\drop{7.5} & 13.7 & 8.9 \\
Macro-only reward & 50.2\drop{2.5} & 51.3 & 7.8  \\
\midrule
\multicolumn{4}{l}{\textit{Ablation on Refinement Mechanism}} \\ 
w/o Refinement   & 49.0\drop{4.9} & 36.7 & 9.4 \\
Reflection-based & 49.5\drop{3.9} & 37.9 & 8.6 \\
\bottomrule
\end{tabular}%
}
\caption{Ablation study of \our{}. The first row represents our full framework \our{}. The subsequent rows show the performance drop when removing or replacing specific components.}
\label{tab:ablation}
\end{table}


\section{Conclusion}
In this work, we introduce \our{}, a framework that resolves the inherent challenges of sparse valid paths and noisy context by shifting exploration to a high-level Plan Space. By decoupling strategic planning from execution grounding, \our{} utilizes Structural Refinement for on-policy repair and Dual-Gating Evaluation to strictly validate both physical executability and strategic alignment. Extensive evaluations on the WebArena benchmark demonstrate that \our{} achieves state-of-the-art performance, significantly surpassing action-level baselines in both task effectiveness and search efficiency. These results empirically validate that abstracting raw execution traces into coherent semantic subplans is a crucial paradigm for scaling autonomous agents to complex, long-horizon web navigation tasks.





\bibliographystyle{named}
\bibliography{ijcai26}

\newpage

\appendix
\section*{Appendix}

\section{Baseline Details}
In this section, we provide comprehensive implementation details for the baselines compared in our evaluation. To strictly assess the performance of \our{}, we benchmark against a diverse set of approaches, categorized into Sequential Agents and Policy-Based Strategies.

\paragraph{Webarena~\citep{zhou2023webarena}} adopts the vanilla agent implementation provided by the official WebArena benchmark. This baseline employs a standard sequential execution paradigm, where the agent generates actions step-by-step based on the current observation and interaction history.

\paragraph{Browsergym~\citep{chezelles2024browsergym}} serves as an environment framework that enables agent-environment interaction through executable Python code. The baseline agent within this framework implements a standard ReAct~\citep{yao2022react} paradigm, alternating between reasoning about the current observation and generating Python scripts to execute atomic actions.

\paragraph{AutoEval~\citep{pan2024autonomous}} introduces an autonomous self-refinement mechanism where the agent evaluates its own generated actions against the instruction constraints. By employing a verify-and-refine loop, this method detects and corrects potential errors before execution, thereby mitigating error propagation in long-horizon tasks.

\paragraph{SteP~\citep{sodhi2023step}} proposes a hierarchical architecture based on "Stacked LLM Policies," which dynamically composes specialized policies to handle complex web tasks. This framework defines the agent's control state as a stack of policy calls, enabling it to decompose high-level objectives into sequences of manageable low-level actions while maintaining context.

\paragraph{AWM~\citep{wang2024agent}} introduces a mechanism to store and retrieve successful execution patterns as reusable workflows. By retrieving relevant historical trajectories based on task similarity, it guides the agent in solving new, unseen tasks, effectively bridging the gap between exploration and exploitation.

\paragraph{CER~\citep{liu2025contextual}} facilitates self-improvement by enabling the agent to learn from both successful and failed interaction histories. It employs a hindsight-like replay mechanism to refine the agent's policy, allowing it to correct previous mistakes and optimize decision-making in complex environments.

\paragraph{AgentOccam~\citep{yang2024agentoccam}} AgentOccam addresses the challenge of information overload in web pages by pruning irrelevant observation data. It filters out non-essential HTML elements and visual clutter based on task relevance, ensuring the agent operates within a clean, high-density context window for more accurate perception.

\paragraph{Search Agent~\citep{koh2024tree}} integrates Best-First Search (BFS) with Large Language Models to enhance decision-making in interactive environments. By treating the web navigation task as a search problem, it iteratively generates and evaluates multiple potential action candidates at each step, allowing the agent to explore diverse execution trajectories before committing to a final path.

\paragraph{WebPilot~\citep{zhang2025webpilot}} adopts a Monte Carlo Tree Search (MCTS) framework to guide autonomous web agents. It functions by building a search tree where nodes represent web states and edges correspond to atomic actions. Through the selection, expansion, simulation, and backpropagation phases, it dynamically balances exploration and exploitation to identify optimal interaction sequences.

\paragraph{Branch-and-Browse~\citep{he2025branch}} introduces a tree-structured browsing framework designed to improve the controllability of web agents. It incorporates an action memory mechanism to track historical interactions and state transitions. This design enables the agent to systematically navigate the DOM state tree, facilitating effective backtracking and coverage of complex web interfaces.

\section{Implementation Details}
\subsection{Browser Environment Setup}
Our experimental infrastructure is built upon the \textbf{BrowserGym} framework~\citep{chezelles2024browsergym}, which provides a unified gymnasium-compliant interface for web agents. We utilize \textbf{Playwright} as the underlying engine to programmatically control the browser and execute atomic actions (e.g., \texttt{click}, \texttt{type}).

To ensure strict reproducibility and eliminate variability caused by network latency or external site updates, we deploy the \textbf{WebArena} benchmark environment locally using isolated Docker containers. This self-contained setup ensures a consistent evaluation state across all experimental runs. A notable exception is applied to map-related tasks: we interface with the live OpenStreetMap service to circumvent functional limitations inherent to the local search functionality of the provided map environment.

To facilitate robust perception, \our{} operates on a multimodal observation space that combines visual and structural information. Specifically, the agent receives two complementary data streams at each time step:

\begin{itemize}
    \item \textbf{Set-of-Marks (SoM):} We employ SoM prompting to enhance the visual grounding capabilities of the Vision-Language Model (VLM). This technique dynamically overlays unique numeric IDs and bounding boxes on all interactable elements within the current viewport. By visually tagging components, SoM bridges the gap between raw pixel inputs and discrete action execution, enabling the agent to target elements via precise identifiers rather than ambiguous coordinate regression.
    
    \item \textbf{Accessibility Tree (AxTree):} Complementing the visual input, we extract and serialize the browser's Accessibility Tree into a structured text format. Unlike the raw HTML DOM, which is often cluttered with decorative tags and scripts, the AxTree filters out non-essential nodes to expose only semantically significant elements (e.g., buttons, links, input fields) along with their functional properties (e.g., roles, names, states). This provides the agent with a clean, logic-centric representation of the page structure.
\end{itemize}

\subsection{Action and Observation Space Optimization}
To further enhance the agent's efficiency and robustness, we adopt the optimization strategies proposed in \textbf{AgentOccam}~\citep{yang2024agentoccam}. Specifically, we refine both the action and observation spaces to mitigate the challenges of information overload and large-scale discrete action selection.

\begin{itemize}
    \item \textbf{Action Space Reduction (Action Pruning):} 
    
    The raw action space in web environments typically includes a vast array of redundant or rarely used operations (e.g., complex mouse drag sequences, generic coordinate clicks, or redundant navigation commands). Exposing this full space often leads to \textit{action hallucination}, where the agent generates syntactically correct but functionally invalid commands. 
    
    Following AgentOccam, we prune the action space by restricting the available API to a concise set of high-frequency primitives (e.g., \texttt{click}, \texttt{type}, \texttt{scroll}, \texttt{goto}). Furthermore, we enforce strict argument constraints: interaction actions are only valid if they reference specific element IDs currently visible in the Set-of-Marks (SoM). This constraints-based approach effectively filters out invalid candidates, ensuring that the MCTS process focuses solely on executable and meaningful interactions.

    \item \textbf{Observation Structure Optimization (Tree Pruning):} 
    Raw Accessibility Trees often contain excessive noise, such as nested generic containers (e.g., meaningless \texttt{<div>} wrappers), invisible elements, or decorative nodes that consume valuable context window without contributing to task reasoning. 
    
    We implement the tree pruning algorithm from AgentOccam to optimize the AxTree structure. This method recursively traverses the DOM tree and filters out nodes that are neither \textit{interactable} nor contain \textit{meaningful textual content}. By removing these "distractor" nodes while preserving the hierarchical ancestor-descendant relationships of the remaining semantic elements, we significantly compress the observation length (token reduction) while increasing the information density, allowing the agent to grasp the page layout more accurately.
\end{itemize}

To ensure a rigorous and fair comparison, we uniformly applied these action and observation space optimizations across all implemented baselines (e.g. Action-MCTS). By standardizing the environment interface, we ensure that the superior performance of \our{} is attributed solely to the proposed plan-space search mechanism rather than engineering improvements in the interaction layer.

\section{Prompts}
In this section, we present the detailed system prompts used in \our{}, corresponding to its four core components: the \textbf{Planner} (Figure~\ref{fig:prompt_subplan}), the \textbf{Operator} (Figure~\ref{fig:prompt_operator}), the \textbf{Evaluator} (Figure~\ref{fig:prompt_micro_eval} and Figure~\ref{fig:prompt_macro_eval}), and the \textbf{Reflector} (Figure~\ref{fig:prompt_reflector}).
For clarity and brevity, dynamic runtime contexts—such as real-time accessibility trees, screenshots, and execution histories—are represented by placeholders (e.g., \texttt{\{AxTree\}}, \texttt{\{Screenshot\}}).
\begin{figure*}[t]
\centering
\begin{tcolorbox}[colback=gray!5!white, colframe=gray!60!black, title=Prompt for Planner]
\small\ttfamily
\textcolor{blue}{[System Message]} \\
You are an expert Planner for an autonomous web agent. Your goal is to propose \{branching\_factor\} candidate subplans for the IMMEDIATE NEXT STEP.

\# CRITICAL: CONTINUITY \& CONTEXT AWARENESS \\
You must analyze the `Subplan History` and the `Current State` to determine the next logical move.

1. IF the last subplan was SUCCESSFUL: \\
\hspace*{1em} - Move forward to the next logical stage. \\
2. IF the last subplan FAILED: \\
\hspace*{1em} - DO NOT simply repeat the failed plan. \\
\hspace*{1em} - You MUST propose a fix, a retry with a different method, or a workaround.

\# Generating Candidates \\
All candidates must focus on the SAME immediate objective but vary in HOW to achieve it: \\
- Variation 1 (Standard): The most direct, common way. \\
- Variation 2 (Granular): Breaking the step down into smaller checks. \\
- Variation 3 (Alternative): Using a different UI element or path.

\# Output Format (JSON)
\begin{verbatim}
{
  "subplans": [
    {
      "thought": "Reasoning based on history...",
      "subplan": "Complete natural language plan."
    },
    ...
  ]
}
\end{verbatim}

\textcolor{blue}{[User Message]} \\
\#\#\# User Instruction \#\#\# \\
\{User Goal Description\}

\#\#\# Subplan History \#\#\# \\
Previously executed subplans: \\
\{List of executed subplans with status: Completed / Not Completed\}

\#\#\# Current Screenshot \#\#\# \\
\{Input Image: Current Page Screenshot with SoM Overlays\}

\#\#\# Current Page Accessibility Tree \#\#\# \\
\{Text: Pruned Accessibility Tree Structure\}

IMPORTANT: Please provide \{branching\_factor\} DIVERSE subplan candidates in JSON format.
\end{tcolorbox}
\caption{The prompt template used by the Planner to generate diverse subplan candidates during the tree search.}
\label{fig:prompt_subplan}
\end{figure*}

\begin{figure*}[t]
\centering
\begin{tcolorbox}[colback=gray!5!white, colframe=gray!60!black, title=Prompt for Operator]
\small\ttfamily
\textcolor{blue}{[System Message]} \\
You are a UI Assistant, your goal is to help the user perform tasks using a web browser. \\
You will be provided with task objective, current step, web page observations, previous plans, and interaction history. You need to issue an action for this step.

\# Output Format \\
\#\#\# Interaction History Summary \#\#\# \\
Emphasize all important details in the INTERACTION HISTORY section.

\#\#\# Observation Highlights \#\#\# \\
List the numerical ids of elements... (e.g., `1321, 52, 756`).

\#\#\# Observation Description \#\#\# \\
Describe information in the current axtree and screenshot...

\#\#\# Reason \#\#\# \\
Provide your rationale for proposing the subsequent action commands here.

\#\#\# Action \#\#\# \\
Only a SINGLE action is allowed in this tag... formatted as 

\# General Tips \\
- You may receive historical thoughts and executed actions as context. \\
- Always take into account the current task, the latest page screenshot, and the history. \\
- If you haven't gotten the final exact answer yet, please do not send message to user.

\# Action Space \\
\{Detailed Description of Action Space: click(), type(), etc.\}

\textcolor{blue}{[User Message]} \\
\#\#\# User Instruction \#\#\# \\
\{User Goal Description\}

\#\#\# Current Subplan to Execute \#\#\# \\
\{Natural Language Subplan generated by Planner\}

\#\#\# Interaction History (Current Subplan) \#\#\# \\
\{List of previously executed atomic actions within this subplan\}

\#\#\# Current Screenshot \#\#\# \\
\{Input Image: Current Page Screenshot with SoM Overlays\}

\# Current page Accessibility Tree \\
\{Text: Pruned Accessibility Tree Structure\}

\end{tcolorbox}
\caption{The prompt template used by the Operator to ground high-level subplans into atomic actions.}
\label{fig:prompt_operator}
\end{figure*}

\begin{figure*}[t]
\centering
\begin{tcolorbox}[colback=gray!5!white, colframe=gray!60!black, title=Prompt for Evaluator(Micro-Scoring)]
\small\ttfamily
\textcolor{blue}{[System Message]} \\
You are a precise evaluator for a web navigation agent. \\
Your ONLY job is to determine if the specific subplan below was successfully executed based on the evidence provided.

**Subplan to Evaluate:** "\{Target Subplan Text\}"

**Evaluation Checklist (Mental Step-by-Step):** \\
1. **Check Errors**: Did the Interaction History show any system errors? If yes -> NO. \\
2. **[CRITICAL] Check Terminal Actions**: \\
- Did the agent execute a completion action like send message to user? \\
\hspace*{1em} - Is the answer VALID? (non-empty, contains info, not a refusal). \\
3. **Check Action Fidelity**: Did the agent actually perform the actions described in the subplan? \\
4. **Check State Change**: Compare the "Pre" and "Post" screenshots. \\
\hspace*{1em} - Is there VISIBLE evidence that the action took effect?

**Format:** \\
Thoughts: <Analyze the delta between Pre and Post states. Point out specific visual changes or errors.> \\
Completed: "yes" or "no"

\textcolor{blue}{[User Message]} \\
\#\#\# Current Subplan \#\#\# \\
\{Target Subplan Text\}

\#\#\# Interaction History (Actions for This Subplan) \#\#\# \\
\{List of atomic actions executed specifically for this subplan\}

\#\#\# Initial State (Before Subplan Execution) \#\#\# \\
\{Image: Screenshot before execution\} \\
\{Text: AxTree snippet before execution\}

\#\#\# Current State (After Subplan Execution) \#\#\# \\
\{Image: Screenshot after execution\} \\
\{Text: AxTree snippet after execution\}

\end{tcolorbox}
\caption{The prompt template used by the Evaluator for the \textbf{Micro-Level Check}.}
\label{fig:prompt_micro_eval}
\end{figure*}

\begin{figure*}[t]
\centering
\begin{tcolorbox}[colback=gray!5!white, colframe=gray!60!black, title=Prompt for Evaluator(Macro-Scoring)]
\small\ttfamily
\textcolor{blue}{[System Message]} \\
You are an expert in evaluating the utility of subplans for completing web navigation tasks. \\
Your goal is to estimate the State Value (V(s)) of the current webpage. Ask yourself: "How close are we to the final goal right now?"

\# CRITICAL PRIORITY: HANDLING TERMINAL SUBPLANS \\
If the subplan involves finishing the task:
\begin{itemize}[leftmargin=*, noitemsep, topsep=0pt]
    \item Completion Signal: Treat the task as functionally completed.
    \item Leniency: If the message is RELEVANT, lean heavily towards Status A.
    \item Minimum Score: Unless completely hallucinated, do NOT rate as C/D/E.
\end{itemize}

\# Evaluation Criteria \\
- Previous Progress: Review subplan history. \\
- Contribution: Assess how much this subplan moved us toward the goal. \\
- Penalties: Error pages (404), backward movement, or repeating actions must be penalized.

\# STATUS CODES (Score Space) \\
A. SUCCESS: Task completed or fully fulfilled requirements. \\
B. ALMOST FINISHED: Extremely close (e.g., 1-2 steps left). \\
C. ON TRACK: Significant progress made, working correctly. \\
D. UNCLEAR: Unsure if positive contribution. \\
E. FAILURE: Stuck, error, or moved backwards.

\# Output Format \\
Thoughts: <Detailed analysis of contribution to global goal> \\
STATUS CODE: A, B, C, D, or E \\
Notes: <Key observations for future steps>

\textcolor{blue}{[User Message]} \\
\#\#\# Overall Task Objective \#\#\# \\
\{User Goal Description\}

\#\#\# Previously Executed Subplans \#\#\# \\
\{History list with previous status codes and scores\}

\#\#\# Initial State (Before Current Subplan) \#\#\# \\
\{Image \& Text: State before execution\}

\#\#\# Current Subplan Action History \#\#\# \\
\{List of atomic actions executed in this step\}

\#\#\# Current Subplan Being Evaluated \#\#\# \\
\{Target Subplan Text\}

\#\#\# Current State (After Subplan Execution) \#\#\# \\
\{Image \& Text: State after execution\}

\end{tcolorbox}
\caption{The prompt template used by the Evaluator for the \textbf{Macro-Level Check}. }
\label{fig:prompt_macro_eval}
\end{figure*}

\begin{figure*}[t]
\centering
\begin{tcolorbox}[colback=gray!5!white, colframe=gray!60!black, title=Prompt for Reflector]
\small\ttfamily
\textcolor{blue}{[System Message]} \\
You are the "Reflector" module of an autonomous web navigation agent. \\
Your job is to analyze a FAILED execution of a subplan and generate a FIXED subplan.

\# Diagnosis Strategy \\
1. TYPE A: Feasibility Error (Wrong Direction)
\begin{itemize}[leftmargin=*, noitemsep, topsep=0pt]
    \item Symptoms: Hallucination, logically blocked path, or impossible goal.
    \item Fix Strategy (Pivot): Propose an ALTERNATIVE approach. Abandon current method.
\end{itemize}
2. TYPE B: Complexity Error (Granularity Issue)
\begin{itemize}[leftmargin=*, noitemsep, topsep=0pt]
    \item Symptoms: Plan is valid but timed out, got stuck, or too many steps.
    \item Fix Strategy (Decompose): Reduce GRANULARITY. Extract ONLY the first, immediate logical segment.
\end{itemize}

\# Goal \\
Generate a REVISED subplan that achieves the SAME goal as the original plan but avoids the previous error.

\# Requirements \\
- The revised plan must be executable in 2-5 concrete actions. \\
- Be specific about element identifiers. \\
- Include necessary preconditions (close popups, wait for load).

\# Output Format (JSON ONLY)
\begin{verbatim}
{
  "reason": "Brief explanation of failure (Type A or B)...",
  "revised_plan": "The new, corrected step-by-step natural language instruction"
}
\end{verbatim}

\textcolor{blue}{[User Message]} \\
\#\# Context \#\# \\
**Original Subplan**: "\{Failed Subplan Text\}"

\#\# Execution Trace \#\# \\
\{List of attempted actions, observations, and specific error messages\}

\#\# Last Screenshot \#\# \\
\{Image: Final state showing where it got stuck\}

\#\# Last Page State (Accessibility Tree) \#\# \\
\{Text: Final AxTree snippet\}

\#\# Task \#\# \\
Analyze the failure reason based on the Trace and Last Observation. \\
Provide a JSON response with the 'reason' and the 'revised plan'.

\end{tcolorbox}
\caption{The prompt template used by the Reflector module to perform \textbf{Structural Refinement}. When a subplan fails execution, this module diagnoses the failure type (Feasibility vs. Complexity) and generates a revised, executable subplan (Pivot or Decompose) to recover the search process.}
\label{fig:prompt_reflector}
\end{figure*}

\end{document}